\documentclass[10pt,twocolumn,letterpaper]{article}

\usepackage{iccv}
\usepackage{times}
\usepackage{epsfig}
\usepackage{graphicx}
\usepackage{amsmath}
\usepackage{amssymb}

\usepackage{subfiles}
\usepackage{cancel}
\usepackage{subfigure}
\usepackage{amsthm}
\usepackage{multirow}
\usepackage{booktabs}
\usepackage{scalerel}
\usepackage{color, colortbl}
\usepackage{threeparttable}
\definecolor{tb_bg_color}{rgb}{0.835, 0.902, 0.941}
\DeclareMathOperator*{\argmax}{arg\,max}

\newtheorem{theorem}{Theorem}

\usepackage[pagebackref=true,breaklinks=true,letterpaper=true,colorlinks,bookmarks=false]{hyperref}

\iccvfinalcopy 

\def\iccvPaperID{7455} 

\ificcvfinal\fi

\begin{document}

\title{FcaNet: Frequency Channel Attention Networks}

\author{Zequn Qin$^1$, Pengyi Zhang$^1$, Fei Wu$^{1,2}$, Xi Li$^{1,2}$\thanks{Corresponding author.}
\\
$^1$College of Computer Science, Zhejiang University,
\\
$^2$Shanghai Institute for Advanced Study, Zhejiang University
\\
{\tt \small zequnqin@gmail.com, pyzhang@zju.edu.cn, wufei@cs.zju.edu.cn, xilizju@zju.edu.cn}
}

\maketitle
\ificcvfinal\thispagestyle{empty}\fi

\begin{abstract}
Attention mechanism, especially channel attention, has gained great success in the computer vision field. Many works focus on how to design efficient channel attention mechanisms while ignoring a fundamental problem, i.e., channel attention mechanism uses scalar to represent channel, which is difficult due to massive information loss. In this work, we start from a different view and regard the channel representation problem as a compression process using frequency analysis. Based on the frequency analysis, we mathematically prove that the conventional global average pooling is a special case of the feature decomposition in the frequency domain. With the proof, we naturally generalize the compression of the channel attention mechanism in the frequency domain and propose our method with multi-spectral channel attention, termed as FcaNet. FcaNet is simple but effective. We can change a few lines of code in the calculation to implement our method within existing channel attention methods. Moreover, the proposed method achieves state-of-the-art results compared with other channel attention methods on image classification, object detection, and instance segmentation tasks. Our method could consistently outperform the baseline SENet, with the same number of parameters and the same computational cost. Our code and models will are publicly available at \url{https://github.com/cfzd/FcaNet}.
\end{abstract}
\section{Introduction}
\label{sec_intro}

As an important and challenging problem in feature modeling, attention mechanisms for convolutional neural networks (CNNs) have recently attracted considerable attention and are widely used in many fields like computer vision \cite{xu2015show} and natural language processing \cite{vaswani2017attention}. In principle, they aim at selectively concentrating on some important information and have many types of variants (e.g., spatial attention, channel attention, and self-attention) corresponding to different feature dimensions. Due to the simplicity and effectiveness in feature modeling, channel attention directly learns to attach importance weights with different channels, becoming a popular and powerful tool for the deep learning community.

\begin{figure}
	\centering
	\includegraphics[width=0.95\linewidth]{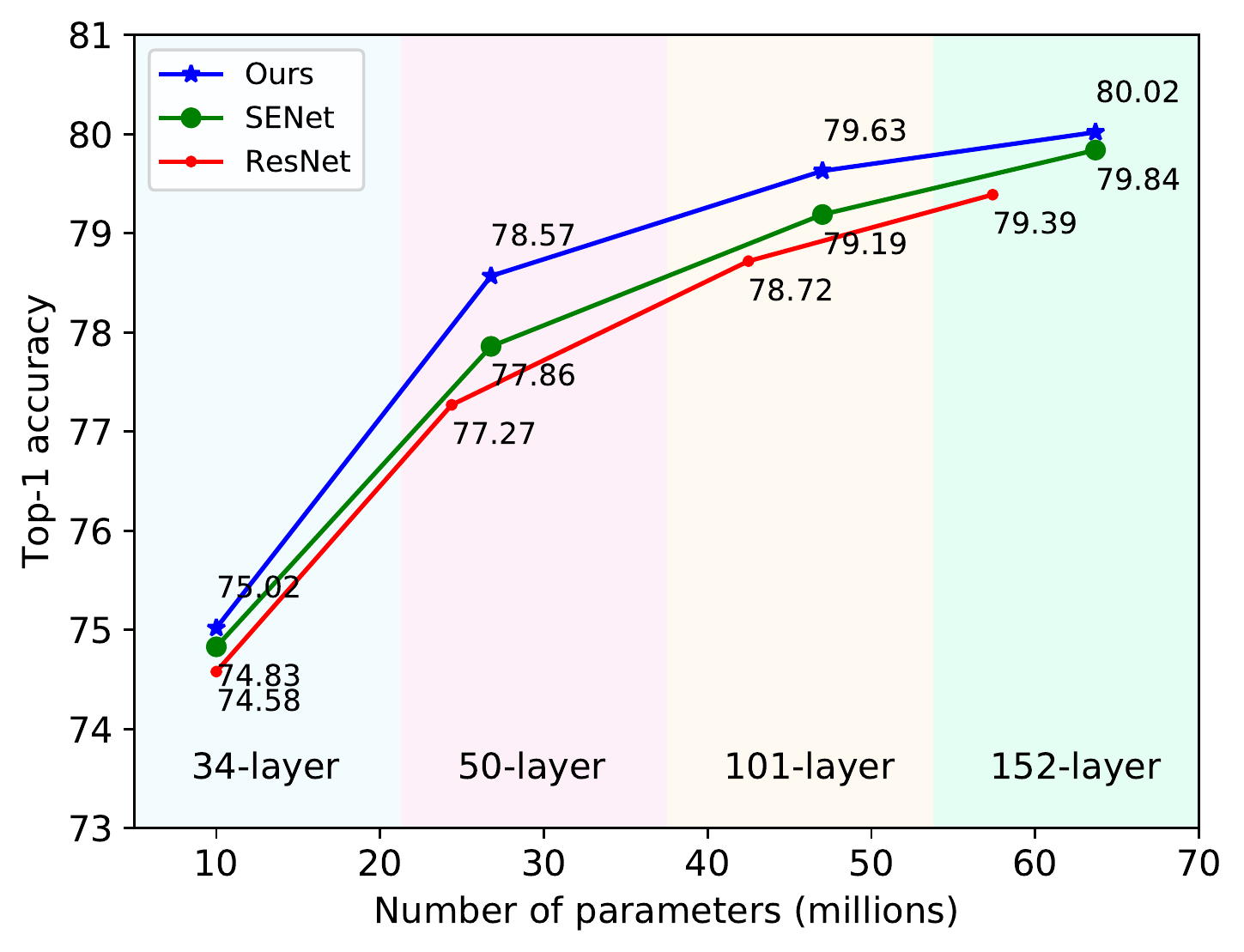}
	\caption{Classification accuracy comparison on ImageNet. With the same number of parameters and computational cost, our method consistently outperforms the baseline SENet. }
	\label{fig_intro}
\end{figure}

Typically, a core step of channel attention approaches is to use a scalar for each channel to conduct the calculation due to the constrained computational overhead, and global average pooling (GAP) becomes the de-facto standard choice in the deep learning community because of its simplicity and efficiency. Nevertheless, every rose has its thorn. The simplicity of GAP makes it hard to well capture complex information for various inputs. Some methods like CBAM~\cite{woo2018cbam} and SRM~\cite{lee2019srm} further use global max pooling and global standard deviation pooling to enhance the performance of GAP. Different from previous works, we consider the scalar representation of a channel as a compression problem. 
Namely, the information of a channel should be compactly encoded by a scalar while preserving the representation ability of the whole channel as much as possible.
In this way, how to effectively compress a channel with a scalar due to the constrained computational overhead is a major difficulty, and it is crucial to channel attention. 

With the above motivation, we propose to use the discrete cosine transform (DCT) to compress channels in the channel attention mechanism for the following reasons: 1) DCT is a widely used data compression method in signal processing, especially with digital images and videos. Many widely used image and video formats like JPEG, HEIF, MPEG, and H.26x use DCT to realize data compression. DCT has a strong energy compaction property~\cite{ahmed1974discrete, rao2014discrete}, so it could achieve high data compression ratio with high quality~\cite{barbero1992dct, lea1994video}. This property meets the demand of the channel attention that representing a channel with a scalar. 2) DCT can be implemented with an element-wise multiplication, and it is differentiable. In this way, it can be easily integrated into CNNs. 3) Surprisingly, DCT can be viewed as a generalization of GAP. Mathematically, GAP (showing the effectiveness in SENet \cite{hu2018squeeze}) is only equivalent to the lowest frequency components of DCT, leaving many other potentially useful frequency components unexplored. This strongly motivates us to tailor DCT for the channel attention mechanism.

In this paper, we further propose a simple, novel, but effective multi-spectral channel attention (MSCA) framework based on the above discussions. In order to better compress channels and explore the components left out by GAP, we propose to tailor DCT and use multiple but limited frequency components of DCT for the channel attention mechanism. 
Note that although we use multi-spectral channel attention, each channel is still represented by only one scalar. Along with the MSCA framework, how to select the frequency component of DCT for each channel is important. In this way, we propose three kinds of frequency component selection criteria to fulfill and validate the MSCA framework, which are LF (Low Frequency based selection), TS (Two-Step selection), and NAS (Neural Architecture Search selection). With these selection criteria, our method achieves state-of-the-art performance against the other channel attention ones.

In a word, the main contribution of this work can be summarized as follows.
\begin{itemize}
	\item We regard the channel attention as a compression problem and introduce DCT in the channel attention. We then prove that conventional GAP is a special case of DCT. Based on this proof, we generalize the channel attention in the frequency domain and propose our method with the multi-spectral channel attention framework, termed as FcaNet.
	\item We propose three kinds of frequency component selection criteria along with the proposed multi-spectral channel attention framework to fulfill FcaNet.
	\item Extensive experiments demonstrate the proposed method achieves state-of-the-art results on both ImageNet and COCO datasets, with the same computational cost as SENet. The results on ImageNet are shown in Fig.~\ref{fig_intro}.
\end{itemize}

\section{Related Work}
\paragraph{Attention Mechanism in CNNs} In \cite{xu2015show}, a visual attention method is first proposed to model the importance of features in the image caption task. Then many methods start to focus on the attention mechanism. A residual attention network \cite{wang2017residual} is proposed with a spatial attention mechanism using downsampling and upsampling. Besides, SENet \cite{hu2018squeeze} proposes the channel attention mechanism. It performs GAP on the channels and then calculates the weights of each channel using fully connected layers. What's more, GE \cite{hu2018gather} uses spatial attention to better exploit the feature context, and $A^2$-Net \cite{chen20182} builds a relation function for image or video recognition. 

Inspired by these works, a series of works like BAM \cite{park2018bam}, DAN \cite{fu2019dual}, CBAM \cite{woo2018cbam}, scSE~\cite{roy2018recalibrating}, and CoordAttention~\cite{hou2021coordinate} are proposed to fuse spatial attention \cite{zhu2019empirical} and channel attention. Among them, CBAM claims that GAP could only get a sub-optimal feature because of the loss of information. For addressing this problem, it uses both the GAP and the global max pooling and gains significant performance improvement. Similarly, SRM~\cite{lee2019srm} also propose to use GAP with global standard deviation pooling. Motivated by CBAM, GSoP \cite{gao2019global} introduces a second-order pooling method for downsampling. NonLocal \cite{wang2018non} proposes to build a dense spatial feature map. AANet \cite{bello2019attention} proposes to embed the attention map with position information into the feature. SkNet \cite{li2019selective} introduces a selective channel aggregation and attention mechanism, and ResNeSt \cite{zhang2020resnest} proposes a similar split attention method. Due to the complicated attention operation, these methods are relatively large. To improve efficiency, GCNet \cite{cao2019gcnet} proposes to use a simple spatial attention module and replace the original spatial downsampling process. ECANet \cite{wang2020eca} introduces one-dimensional convolution layers to reduce the redundancy of fully connected layers and obtains more efficient results. 

Besides these works, many methods try to extend the attention mechanism to specific tasks, like multi-label classification \cite{guo2019visual}, saliency detection \cite{zhao2019pyramid}, visual explanation \cite{fukui2019attention}, and super-resolution \cite{zhang2018image}.

\paragraph{Frequency Domain Learning} 
Frequency analysis has always been a powerful tool in the signal processing field. In recent years, some applications of introducing frequency analysis in the deep learning field emerge. In \cite{ehrlich2019deep, gueguen2018faster}, frequency analysis is introduced in the CNNs by JPEG encoding. Then, DCT is incorporated in \cite{xu2020learning} to reduce communication bandwidth. There are also some applications in the model compression and pruning tasks like \cite{chen2016compressing, liu2018frequency, wang2016cnnpack}.

\section{Method}

In this section, we first revisit the formulation of DCT and channel attention. Then, based on these works, we elaborate on the derivation of our multi-spectral channel attention framework. Meanwhile, along with the multi-spectral channel attention framework, three kinds of frequency components selection methods are proposed.

\subsection{Revisiting DCT and Channel Attention}

We first elaborate on the definitions of discrete cosine transform and channel attention mechanism. 

\label{sec_ca_and_dct}

\paragraph{Discrete Cosine Transform (DCT)}

Typically, the basis function of two-dimensional (2D) DCT \cite{ahmed1974discrete} is:
\begin{equation}
    B_{h,w}^{i,j} = \cos(\dfrac{\pi h}{H}(i+\dfrac{1}{2}))\cos(\dfrac{\pi w}{W}(j+\dfrac{1}{2})).
    \label{eq_basis}
\end{equation}
Then the 2D DCT can be written as:
\begin{equation}
    \begin{aligned}
        &f_{h,w}^{2d}  =  \sum_{i=0}^{H-1} \sum_{j=0}^{W-1} x_{i,j}^{2d} B_{h,w}^{i,j}\\
        s.t. \;\; h \in \{0,1, & \cdots,H-1\}, w \in \{0,1,\cdots,W-1\},
    \end{aligned}
    \label{eq_2ddct}
\end{equation}
in which $f^{2d} \in \mathbb{R}^{H \times W}$ is the 2D DCT frequency spectrum, $x^{2d} \in \mathbb{R}^{H \times W}$ is the input, $H$ is the height of $x^{2d}$, and $W$ is the width of $x^{2d}$. Correspondingly, the inverse 2D DCT can be written as:
\begin{equation}
    \begin{aligned}
        & x_{i,j}^{2d} =  \sum_{h=0}^{H-1} \sum_{w=0}^{W-1} f_{h,w}^{2d} B_{h,w}^{i,j},\\
         s.t. \;\; i \in \{0,1, & \cdots,H-1\}, j \in \{0,1,\cdots,W-1\}.
    \end{aligned}
    \label{eq_2didct}
\end{equation}
Please note that in Eqs.~\ref{eq_2ddct} and \ref{eq_2didct}, some constant normalization factors are removed for simplicity, which will not affect the results in this work.

\paragraph{Channel Attention}
The channel attention mechanism is widely used in CNNs. It uses scalar to represent and evaluate the importance of each channel. Suppose $X \in \mathbb{R}^{C \times H \times W}$ is the image feature tensor in networks, $C$ is the number of channels, $H$ is the height of the feature, and $W$ is the width of the feature. As discussed in Sec. \ref{sec_intro}, we treat the scalar representation in channel attention as a compression problem since it has to represent the whole channel while only one scalar can be used. In this way, the attention mechanism can be written as:
\begin{equation}
    att = sigmoid(fc(compress(X))),
    \label{eq_ca}
\end{equation}
where $att \in \mathbb{R} ^ {C}$ is the attention vector, $sigmoid$ is the Sigmoid function, $fc$ represents the mapping functions like fully connected layer or one-dimensional convolution, and $compress: \mathbb{R}^{C \times H \times W} \mapsto \mathbb{R}^C $ is a compression method. After obtaining the attention vector of all $C$ channels, each channel of input $X$ is scaled by the corresponding attention value:
\begin{equation}
    \widetilde{X}_{:,i,:,:} = att_i X_{:,i,:,:}, \:\:\: s.t. \;\; 
    i \in \{0,1,\cdots,C-1\},
\end{equation}
in which $\widetilde{X}$ is the output of attention mechanism, $att_i$ is the $i$-th element of attention vector, and $X_{:,i,:,:}$ is the $i$-th channel of input.

Typically, global average pooling is the de-facto compression method \cite{hu2018squeeze,wang2020eca} for its simplicity and effectiveness. There are also compression methods like global max pooling~\cite{woo2018cbam} and global standard deviation pooling~\cite{lee2019srm}. 



\subsection{Multi-Spectral Channel Attention}
\label{sec_msca}
In this section, we first theoretically discuss the problem of existing channel attention mechanisms. Based on the theoretical analysis, we then elaborate on the network design of the proposed method.

\paragraph{Theoretical Analysis of Channel Attention}

As discussed in Sec.~\ref{sec_ca_and_dct}, DCT can be viewed as a weighted sum of inputs. We further prove that GAP is actually a special case of 2D DCT.
\begin{theorem}
    GAP is a special case of 2D DCT, and its result is proportional to the lowest frequency component of 2D DCT.
    \label{theo}
\end{theorem}

\begin{figure*}[t]
    \centering
    \subfigure[Original SENet]{\includegraphics[width=0.93\textwidth]{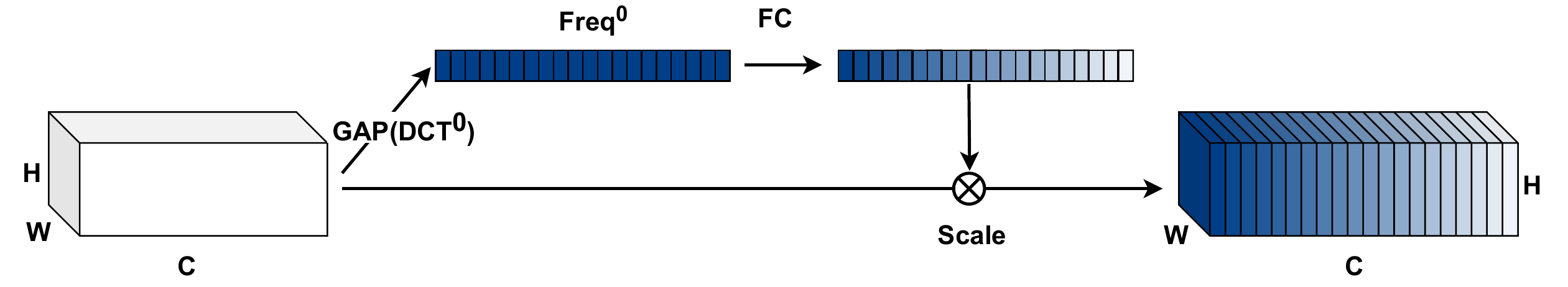}}
    \subfigure[Multi-spectral channel attention]{\includegraphics[width=0.93\textwidth]{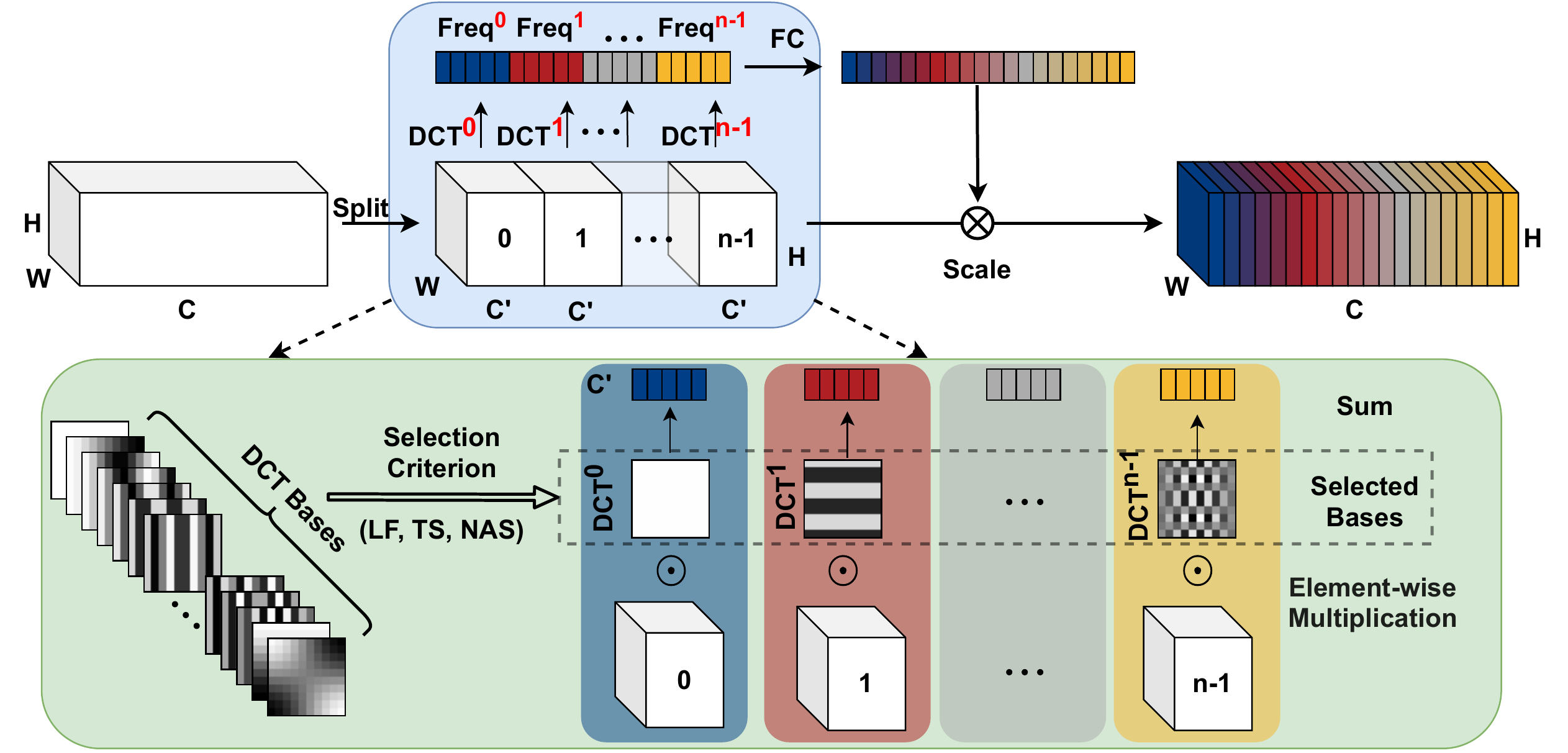}}
    \caption{Illustration of existing channel attention and multi-spectral channel attention. For simplicity, the 2D DCT indices are represented in the one-dimensional format. We can see that our method uses multiple frequency components with the selected DCT bases, while SENet only uses GAP in channel attention. Best viewed in color.}
    \label{fig_main}
\end{figure*}

\begin{proof} Suppose $h$ and $w$ in Eq.~\ref{eq_2ddct} are $0$, we have:
\begin{equation}
    \begin{aligned}
        f_{\textcolor{red}{0,0}}^{2d} &= \sum_{i=0}^{H-1} \sum_{j=0}^{W-1} x_{i,j}^{2d} \cos(\dfrac{\textcolor{red}{0}}{H}(i+\dfrac{1}{2}))\cos(\dfrac{\textcolor{red}{0}}{W}(j+\dfrac{1}{2}))   \\
        & = \sum_{i=0}^{H-1} \sum_{j=0}^{W-1} x_{i,j}^{2d} \\
        & = gap(x^{2d}) HW.
    \end{aligned}
    \label{eq_proof}
\end{equation}

In Eq.~\ref{eq_proof}, $f_{0,0}^{2d}$ represents the lowest frequency component of 2D DCT, and it is proportional to GAP. In this way, Theorem \ref{theo} is proved.
\end{proof}

\paragraph{Multi-Spectral Channel Attention Module}
Based on the theoretical analysis and Theorem \ref{theo}, we can see that using GAP in the channel attention mechanism means only the lowest frequency information is preserved. And all components from other frequencies are discarded, which also encode the useful information patterns in representing the channels and should not be left out. 

To better compress channels and introduce more information, we propose to generalize GAP to more frequency components of 2D DCT and compress more information with multiple frequency components of 2D DCT, including the lowest frequency component, i.e., GAP.

First, the input $X$ is split into many parts along the channel dimension. Denote $[X^0,X^1, \cdots, X^{n-1}]$ as the parts, in which $X^i \in \mathbb{R}^{C' \times H \times W}$, $i \in \{0,1, \cdots, n-1\}$, $C' = \dfrac{C}{n}$, and $C$ should be divisible by $n$. For each part, a corresponding 2D DCT frequency component is assigned, and the 2D DCT results can be used as the compression results of channel attention. In this way, we have:
\vspace{-2pt}
\begin{equation}
    \begin{aligned}
        Freq^i & = \text{2DDCT}^{u_i,v_i}(X^i),  \\
        & = \sum_{h=0}^{H-1} \sum_{w=0}^{W-1} X_{:,h,w}^i B_{h,w}^{u_i,v_i} \\
        & s.t. \;\; i \in \{0,1,\cdots,n-1\},
    \end{aligned}
    \label{eq_2ddct_network}
\end{equation}
in which $[u_i,v_i]$ are the frequency component 2D indices corresponding to $X^i$, and $Freq^i \in \mathbb{R}^{C'}$ is the $C'$-dimensional vector after the compression. The whole compression vector can be obtained by concatenation:
\vspace{-2pt}
\begin{equation}
    \begin{aligned}
    Freq = & compress(X)\\
        = & \text{cat}([Freq^0,Freq^1,\cdots,Freq^{n-1}]),
    \end{aligned}
    \label{eq_cat}
\end{equation}
in which $Freq \in \mathbb{R}^C$ is the obtained multi-spectral vector.
The whole multi-spectral channel attention framework can be written as:
\begin{equation}
    ms\_att = sigmoid(fc(Freq)).
    \label{eq_final}
\end{equation}

From Eqs.~\ref{eq_cat} and \ref{eq_final}, we can see that our method generalizes the original GAP method to a framework with multiple frequency components. By doing so, the channel information after compression is effectively enriched for representation. The overall illustration of our method is shown in Fig.~\ref{fig_main}.

\paragraph{Criteria for Choosing Frequency Components}
There exists an important problem of how to choose frequency component indices $[u_i,v_i]$ for each part $X^i$. In order to fulfill the multi-spectral channel attention, we propose three kinds of criteria, which are FcaNet-LF (Low Frequency), FcaNet-TS (Two-Step selection), and FcaNet-NAS (Neural Architecture Search). 

\textbf{FcaNet-LF} means FcaNet with low-frequency components. As we all know, many compression methods use low-frequency information of DCT to compress information. Moreover, some methods \cite{hu2018squeeze, xu2020learning} have shown CNNs prefer low-frequency information. In this way, the first criterion for choosing frequency components is to only select low-frequency components.

\textbf{FcaNet-TS} means FcaNet selects components within a two-step selection scheme. Its main idea is to first determine the importance of each frequency component and then investigate the effects of using different numbers of frequency components. Namely, we evaluate the results of each frequency component in channel attention individually. Finally, we choose the Top-k highest performance frequency components based on the evaluation results.

\textbf{FcaNet-NAS} means FcaNet with searched components. For this criterion, we use neural architecture search to search the best frequency component for channels. For each part $X^i$, a set of continuous variables $\alpha=\{\alpha^{(u,v)}\}$ are assigned to search components. The frequency components of this part can be written as:
\begin{equation}
    Freq^i_{nas} = \sum_{\scaleto{(u, v) \in O}{6pt}} \dfrac{\exp(\alpha^{(u,v)})}{ \sum_{\scaleto{(u', v') \in O}{6pt}  } \exp(\alpha^{(u',v')}) } \text{2DDCT}^{u,v}(X^i),
\end{equation}
in which $O$ is the set containing all 2D DCT frequency component indices.
After training, the frequency component for $X^i$ is derived by 
$(u_i^*,v_i^*) = \argmax_{(u,v) \in O} \{\alpha^{(u,v)}\}$.

The ablation studies about these criteria can be seen in Sec.~\ref{sec_ablation}.

\section{Experiments}
In this section, we first elaborate on the details of our experiments. Second, we show the ablation studies about FcaNet. Third, we give discussions about how the information is compressed in our framework, complexity, and code implementation. At last, we investigate the effectiveness of our method on the task of image classification, object detection, and instance segmentation. 

\subsection{Implementation Details}
\label{Implementation Details}
To evaluate the results of the proposed FcaNet on ImageNet~\cite{ILSVRC15}, we employ four widely used CNNs as backbone models, including ResNet-34, ResNet-50, ResNet-101, and ResNet-152. We follow the data augmentation and hyper-parameter settings in~\cite{he2016deep} and~\cite{he2019bag}. Concretely, the input images are cropped randomly to 224$\times$224 with random horizontal flipping. We use an SGD optimizer with a momentum of 0.9, a weight decay of 1e-4, and a batch size of 128 per GPU at training time. For large models like ResNet-101 and ResNet-152, the batch size is set to 64. The learning rate is set to 0.1 for a batch size of 256 with the linear scaling rule~\cite{goyal2017accurate}.
All models are trained within 100 epochs with cosine learning rate decay and label smoothing. Notably, for training efficiency, we use the Nvidia APEX mixed precision training toolkit.

To evaluate our method on MS COCO~\cite{lin2014microsoft} using Faster R-CNN~\cite{ren2015faster} and Mask R-CNN~\cite{he2017mask}. We use the implementation of detectors from the MMDetection~\cite{mmdetection} toolkit and employ its default settings. During training, the shorter side of the input image is resized to 800. All models are optimized using SGD with a weight decay of 1e-4, a momentum of 0.9, and a batch size of 2 per GPU within 12 epochs. The learning rate is initialized to 0.01 and is decreased by the factor of 10 at the 8th and 11th epochs, respectively.

All models are implemented in PyTorch~\cite{paszke2019pytorch} framework and with eight Nvidia RTX 2080Ti GPUs.

\subsection{Ablation Study}
\label{sec_ablation}

As discussed in Sec.~\ref{sec_msca}, we propose three kinds of criteria, including FcaNet-LF (Low Frequency), FcaNet-TS (Two-Step selection), and FcaNet-NAS (Neural Architecture Search). In this section, we first show the ablations about these variants. Then we discuss the relation between FcaNet and fully learnable channel attention.

\paragraph{The effects of individual frequency components} For FcaNet-TS, the first step is to determine the importance of each frequency component.
To investigate the effects of different frequency components individually in channel attention, we only use one frequency component at a time. We divide the whole 2D DCT frequency space into $7\times7$ parts since the smallest feature map size is $7\times7$ on ImageNet. In this way, there are in total of 49 experiments. To speed up the experiments, we first train a standard ResNet-50 network for 100 epochs as the base model. Then we add channel attention to the base model with different frequency components to verify the effects\footnote{A new version of Fig.3 without any validation data will be updated in the next version soon.}. All added models are trained within 20 epochs with a similar optimization setting in Sec.~\ref{Implementation Details}, while the learning rate is set to 0.02.

\begin{figure}[htbp]
	\centering
	\includegraphics[width=.9\linewidth]{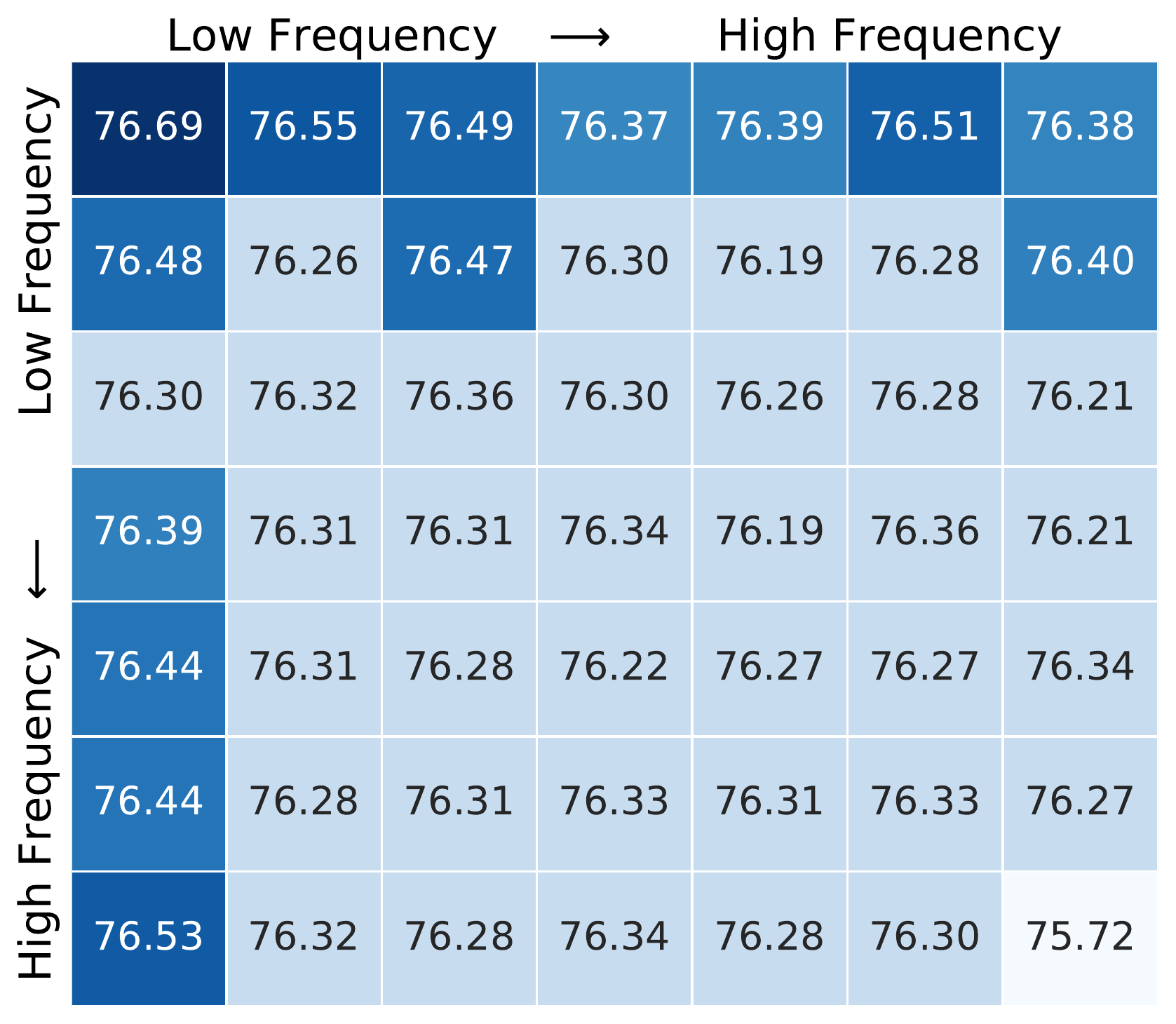}
	\caption{Top-1 accuracies on ImageNet using different frequency components in channel attention individually.}
	\label{fig_frequency_components}
\end{figure}

As shown in Fig.~\ref{fig_frequency_components}, we can see that using lower frequency could have better performance, which is intuitive and verifies the success of SENet. This also verifies the conclusion \cite{xu2020learning} that deep networks prefer low-frequency information. Nevertheless, interestingly, we can see that nearly all frequency components (except the highest component) have very small gaps ($<=0.5\%$ Top-1 accuracy) between the lowest one, i.e., vanilla channel attention with GAP. This shows that other frequency components can also cope well with the channel attention mechanism, and it is effective to generalize the channel attention in the frequency domain.

\paragraph{The effects of different numbers of frequency components}

For FcaNet-LF, we verify the results of using K lowest-frequency components. For FcaNet-TS, we select Top-K highest performance frequency components in Fig. \ref{fig_frequency_components}. For simplicity, K could be 1, 2, 4, 8, 16, or 32.

\begin{figure}[htbp]
	\centering
	\includegraphics[width=.9\linewidth]{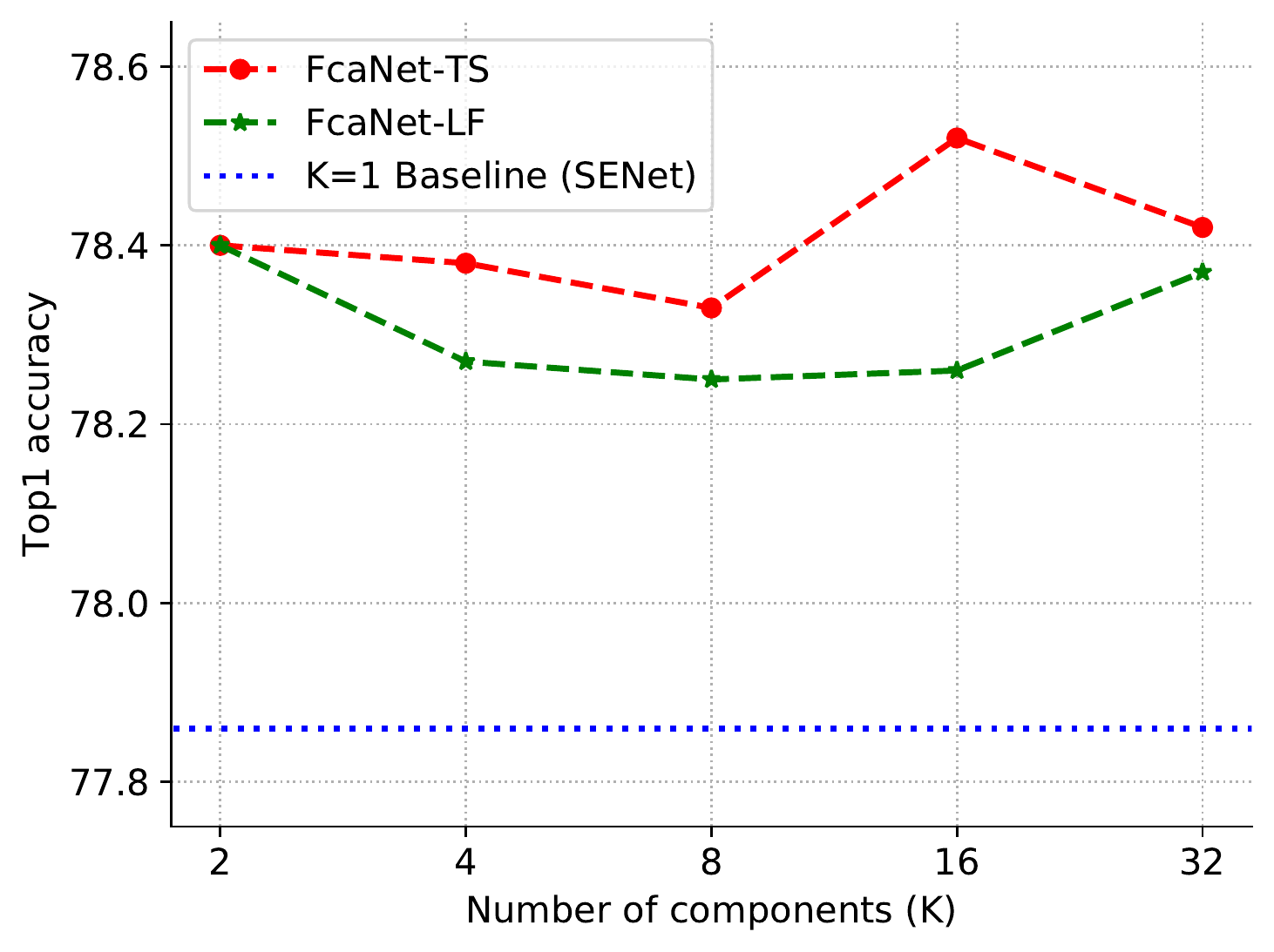}
	\caption{Top1 accuracy with different numbers of components. Since FcaNet-NAS searches and determines frequency components automatically, it is not included in this experiment.}
	\label{fig_num_freq}
\end{figure}
As shown in Fig.~\ref{fig_num_freq}, we can see two phenomena. 1) All experiments with multi-spectral attention have a significant performance gain compared with the one only using the GAP in channel attention. This verifies our idea of using multiple frequency components in channel attention. 2) For FcaNet-LF and FcaNet-TS, the settings with 2 and 16 frequency components gain the best performance, respectively. In this way, we use these settings in our method and all other experiments. 

\paragraph{Comparison with fully learnable channel attention}

As shown in Eq. \ref{eq_2ddct_network}, we use the 2D DCT basis functions to compress channels. The 2D DCT basis functions $B_{h,w}^{u_i,v_i}$ can be simply regarded as a tensor containing DCT coefficients. In this way, a natural question is that how about directly learning a tensor to compress channels. We compare our method with three different kinds of tensors, which are Fixed tensor with Random initialization (FR), Learned tensor with Random initialization (LR), and Learned tensor with DCT initialization (LD). In this case, our method can be viewed as a Fixed tensor with DCT initialization (FD).

\begin{figure}[htbp]
	\centering
	\includegraphics[width=.9\linewidth]{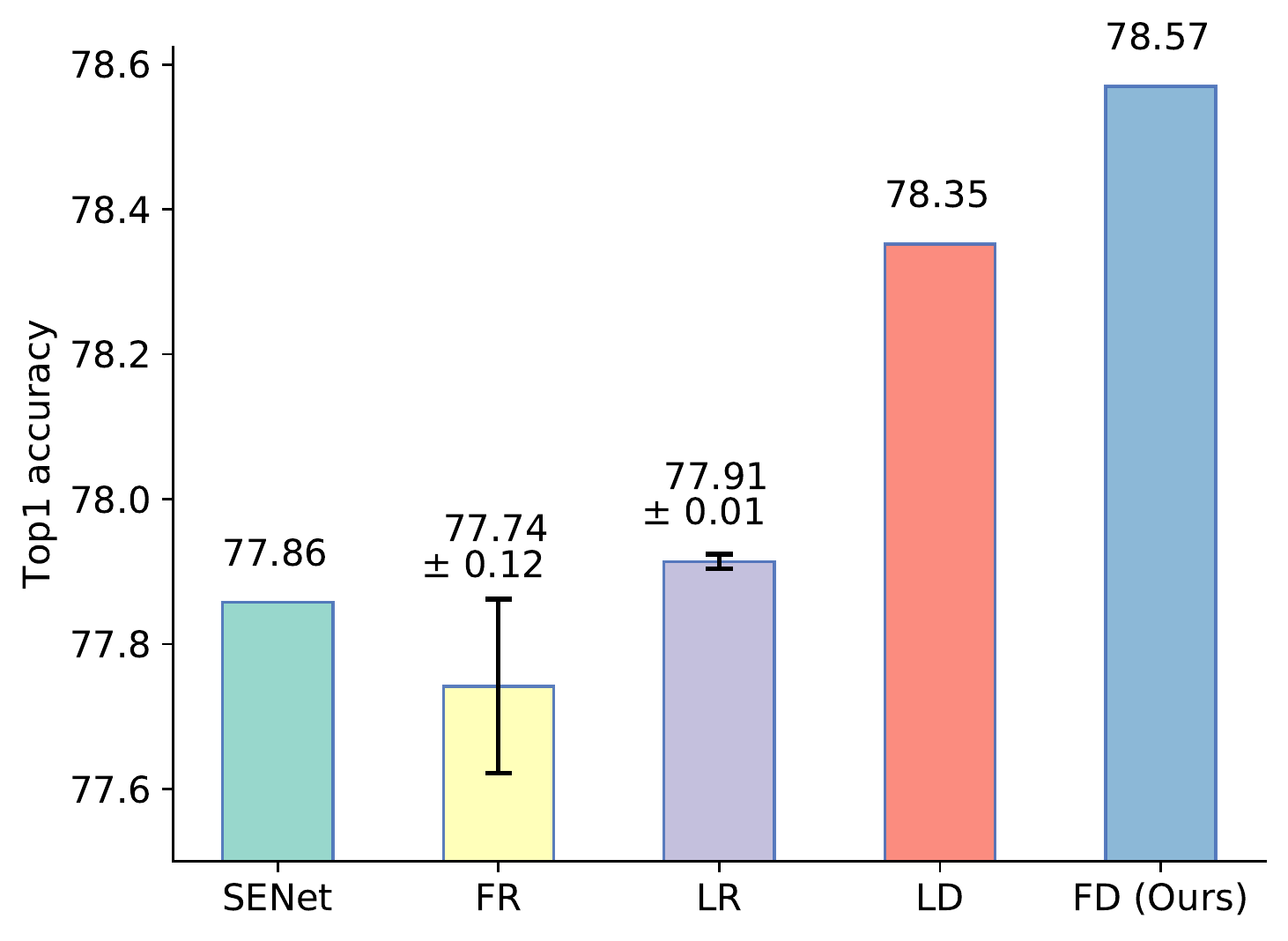}
	\caption{Comparison with fully learnable channel attention. FR means a Fixed tensor with Random initialization, LR means a Learned tensor with Random initialization, LD means a Learned tensor with DCT initialization, and FD means a Fixed tensor with DCT initialization, which is our method. For settings with the random initialization, the error bar is shown. }
	\label{fig_fully_learnable}
\end{figure}

The comparison is shown in Fig. \ref{fig_fully_learnable}. We can see that all settings with the DCT initialization (LD, FD) outperform the ones without DCT (FR, LR). Moreover, the setting with fixed DCT initialization (ours) even outperforms the fully learnable channel attention methods (LR, LD), which shows the effectiveness of using DCT to compress channels.

\subsection{Discussion}

\paragraph{How the multi-spectral framework compresses and embeds more information}
In Sec.~\ref{sec_msca}, we show that only using GAP in channel attention is actually discarding information of all other frequency components except the lowest one, i.e., GAP. In this way, generalizing channel attention in the frequency domain and using the multi-spectral framework could naturally embed more information in the channel attention mechanism. 



\begin{table*}[htbp]
	\centering
	\caption{Comparison of different attention methods on ImageNet. All results are reproduced and trained with the same training setting except AANet, which has no official code. }
	\label{classification}
	\begin{threeparttable}                                        
\begin{tabular}{lcccccccc}
	\toprule
	Method & Years & Backbone & Parameters & FLOPS & Train FPS & Test FPS & Top-1 acc & Top-5 acc \\
	\hline
	ResNet~\cite{he2016deep} & CVPR16 & \multirow{5}{*}{ResNet-34} & 21.80 M & 3.68 G & 2898 & 3840 & 74.58 & 92.05 \\
	SENet~\cite{hu2018squeeze} & CVPR18 &  & 21.95 M & 3.68 G & 2729 & 3489 & 74.83 & 92.23 \\
	ECANet~\cite{wang2020eca} & CVPR20  &  & 21.80 M & 3.68 G & 2703 & 3682 &74.65&92.21\\ 
	\rowcolor{tb_bg_color}
	FcaNet-LF  & &  & 21.95 M  & 3.68 G & 2717 & 3356 & \textbf{74.95} & 92.16\\
	\rowcolor{tb_bg_color}
	FcaNet-TS  & &  & 21.95 M  & 3.68 G & 2717 & 3356 & \textbf{75.02} & 92.07\\
	\rowcolor{tb_bg_color}
	FcaNet-NAS  & &  & 21.95 M  & 3.68 G & 2717 & 3356 & \textbf{74.97} & 92.34 \\  
	\hline
	ResNet~\cite{he2016deep} & CVPR16 & \multirow{10}{*}{ResNet-50} & 25.56 M & 4.12 G & 1644 & 3622 & 77.27 & 93.52\\
	SENet~\cite{hu2018squeeze} & CVPR18 &  & 28.07 M & 4.13 G & 1457 & 3417 &77.86 & 93.87\\
	CBAM~\cite{woo2018cbam}& ECCV18 &  & 28.07 M & 4.14 G &1132 &3319 & 78.24& 93.81\\
	GSoPNet1\tnote{*}~\cite{gao2019global} & CVPR19 &  & 28.29 M & 6.41 G &1095 & 3029& 79.01 & 94.35\\
	GCNet~\cite{cao2019gcnet} & ICCVW19 &  & 28.11 M & 4.13 G &1477&3315& 77.70 & 93.66 \\ 
	AANet~\cite{bello2019attention} & ICCV19 & & 25.80 M & 4.15 G &-&-&77.70 & 93.80\\
	ECANet~\cite{wang2020eca} & CVPR20 &  & 25.56 M & 4.13 G & 1468& 3435& 77.99 & 93.85\\ 
	\rowcolor{tb_bg_color}
	FcaNet-LF &  &  & 28.07 M & 4.13 G &1430&3331& \textbf{78.43} & 94.15\\
	\rowcolor{tb_bg_color}
	FcaNet-TS  &  &  & 28.07 M & 4.13 G &1430&3331& \textbf{78.57} & 94.10\\
	\rowcolor{tb_bg_color}
	FcaNet-NAS &  &  & 28.07 M & 4.13 G &1430&3331& \textbf{78.46} & 94.09\\
	\hline
	ResNet~\cite{he2016deep}& CVPR16 & \multirow{7}{*}{ResNet-101} & 44.55 M & 7.85 G & 816&3187& 78.72 & 94.30 \\
	SENet~\cite{hu2018squeeze}& CVPR18 &  & 49.29 M & 7.86 G &716&2944& 79.19 & 94.50\\
	AANet~\cite{bello2019attention} & ICCV19 & & 45.40 M & 8.05 G &-&-&78.70& 94.40 \\
	ECANet~\cite{wang2020eca} & CVPR20 &  & 44.55 M & 7.86 G &721&3000& 79.09 & 94.38\\ 
	\rowcolor{tb_bg_color}
	FcaNet-LF & &  & 49.29 M & 7.86 G &705&2936& \textbf{79.46} & 94.60\\
	\rowcolor{tb_bg_color}
	FcaNet-TS  & &  & 49.29 M & 7.86 G &705&2936& \textbf{79.63} & 94.63\\
	\rowcolor{tb_bg_color}
	FcaNet-NAS & &  & 49.29 M & 7.86 G &705&2936& \textbf{79.53} & 94.64\\
	\hline
	ResNet~\cite{he2016deep}& CVPR16 & \multirow{7}{*}{ResNet-152} & 60.19 M & 11.58 G &559&2721& 79.39 & 94.74\\
	SENet~\cite{hu2018squeeze}& CVPR18 &  & 66.77 M & 11.60 G &508&2566& 79.84 & 94.82 \\
	AANet~\cite{bello2019attention} & ICCV19 & & 61.60 M & 11.90 G &-&-& 79.10 & 94.60 \\
	ECANet~\cite{wang2020eca}& CVPR20 &  & 60.19 M & 11.59 G &515&2619& 79.86 & 94.80\\ 
	\rowcolor{tb_bg_color}
	FcaNet-LF & &  & 66.77 M & 11.60 G &502&2387& \textbf{80.13} & 94.90 \\
	\rowcolor{tb_bg_color}
	FcaNet-TS  & &  & 66.77 M & 11.60 G &502&2387& \textbf{80.02} & 94.89\\
	\rowcolor{tb_bg_color}
	FcaNet-NAS & &  & 66.77 M & 11.60 G &502&2387& \textbf{79.96} & 94.94 \\
	\bottomrule
	\end{tabular}
	\begin{tablenotes}\footnotesize
		\item[*] Please note that although GSoPNet1 has higher performance, the computational cost is 1.5 times larger than ours. Our method still outperforms GSoPNet1 when compared with the same computational cost. A detailed comparison can be found in the supplementary files.
		\end{tablenotes}
\end{threeparttable}
\end{table*}

Besides the above derivation, we also give a thought experiment to show that more information could be embedded. As we all know, deep networks are redundant \cite{he2017channel,zhuang2018discrimination}. If two channels are redundant for each other, we can only get the same information using GAP. However, in our multi-spectral framework, it is possible to extract more information from redundant channels because different frequency components contain different information. In this way, the proposed multi-spectral framework could embed more information in the channel attention mechanism.

\paragraph{Complexity analysis}
We analyze the complexity of our method from two aspects: the number of parameters and the computational cost.

For the number of parameters, our method has no extra parameters compared with the baseline SENet because the weights of 2D DCT are pre-computed constant.

For the computational cost, our method has a negligible extra cost and can be viewed as having the same computational cost as SENet. With ResNet-34, ResNet-50, ResNet-101, and ResNet-152 backbone, the relative computational cost increases of our method are $0.04\%$, $0.13\%$, $0.11\%$, and $0.11\%$ compared with SENet, respectively. More results can be found in Table~\ref{classification}.

\paragraph{A Few lines of code change}
Another important property of the proposed multi-spectral framework is that it can be easily carried out with existing channel attention implementations. The only difference between our method and SENet is the channel compression method (GAP vs. multi-spectral 2D DCT). As described in Sec.~\ref{sec_ca_and_dct} and Eq.~\ref{eq_2ddct_network}, 2D DCT can be viewed as a weighted sum of inputs. It can be simply implemented by element-wise multiplication and summation. In this way, our method could be easily integrated into arbitrary channel attention methods.

\begin{table*}[htbp]
	\centering
	\caption{Object detection results of different methods on COCO val 2017.}
	\label{detection}
	\begin{tabular}{lccccccccc}
		\toprule
		Method  & Detector & Parameters & FLOPs & AP & $AP_{50}$ & $AP_{75}$ & $AP_{S}$ & $AP_{M}$ &$AP_{L}$\\
		\hline
		ResNet-50 & \multirow{5}{*}{Faster-RCNN} & 41.53 M  & 215.51 G &  36.4   & 58.2 & 39.2 & 21.8 & 40.0 & 46.2\\
		SENet &  & 44.02 M & 215.63 G & 37.7 & 60.1 & 40.9 & 22.9 & 41.9 & 48.2\\
		ECANet & & 41.53 M & 215.63 G & 38.0 & 60.6 & 40.9 & 23.4 & 42.1 & 48.0\\
		\rowcolor{tb_bg_color}
		FcaNet-LF  &  & 44.02 M & 215.63 G & \textbf{39.0} & 61.3 & 41.9 & 23.4 & 42.5 & 49.7\\
		\rowcolor{tb_bg_color}
		FcaNet-TS  &  & 44.02 M & 215.63 G & \textbf{39.0} & 61.1 & 42.3 & 23.7 & 42.8 & 49.6\\
		\rowcolor{tb_bg_color}
		FcaNet-NAS  &  & 44.02 M & 215.63 G & \textbf{39.0} & 60.9 & 42.3 & 23.0 & 42.9 & 49.9\\ \hline
		ResNet-101 & \multirow{5}{*}{Faster-RCNN} & 60.52 M & 295.39 G &  38.7 & 60.6 &41.9 &22.7 &43.2 &50.4\\
		SENet &  & 65.24 M & 295.58 G & 39.6 &62.0 &43.1 &23.7 &44.0 &51.4\\
		ECANet & & 60.52 M & 295.58 G & 40.3 & 62.9 & 44.0 & 24.5 & 44.7 &51.3\\
		\rowcolor{tb_bg_color}
		FcaNet-LF  &  & 65.24 M & 295.58 G & \textbf{41.3} & 63.4 & 44.9 & 24.6 & 45.6 & 53.6\\
		\rowcolor{tb_bg_color}
		FcaNet-TS  &  & 65.24 M & 295.58 G & \textbf{41.2} & 63.3 & 44.6 & 23.8 & 45.2 & 53.1\\
		\rowcolor{tb_bg_color}
		FcaNet-NAS  &  & 65.24 M & 295.58 G & \textbf{41.2} & 63.3 & 44.9 & 24.7 & 45.2 & 53.0\\
		\hline
		ResNet-50 & \multirow{7}{*}{Mask-RCNN} & 44.17 M & 261.81 G & 37.2  & 58.9 & 40.3 & 22.2 & 40.7 & 48.0\\
		SENet &  & 46.66 M & 261.93 G & 38.7 & 60.9 & 42.1 & 23.4 & 42.7 & 50.0\\
		GCNet & & 46.69 M & 261.94 G & 39.4 & 61.6 & 42.4 & - & - & -\\
		ECANet & & 44.17 M & 261.93 G & 39.0 & 61.3 & 42.1 & 24.2 & 42.8 & 49.9\\
		\rowcolor{tb_bg_color}
		FcaNet-LF  &  & 46.66 M & 261.93 G & \textbf{40.3} & 61.9 & 43.9 & 24.9 & 43.6 & 52.2\\
		\rowcolor{tb_bg_color}
		FcaNet-TS  &  & 46.66 M & 261.93 G & \textbf{40.3} & 62.0 & 44.1 & 25.2 & 43.9 & 52.0\\
		\rowcolor{tb_bg_color}
		FcaNet-NAS  &  & 46.66 M & 261.93 G & \textbf{40.3} & 61.9 & 43.9 & 24.9 & 43.6 & 52.2\\
		\bottomrule
	\end{tabular}
\end{table*}

\subsection{Image Classification on ImageNet}
We compare our FcaNet with the state-of-the-art methods using ResNet-34, ResNet-50, ResNet-101, and ResNet-152 backbones on ImageNet, including SENet~\cite{hu2018squeeze}, CBAM~\cite{woo2018cbam}, GSoP-Net1~\cite{gao2019global}, GCNet~\cite{cao2019gcnet}, AANet~\cite{bello2019attention}, and ECANet~\cite{wang2020eca}. The evaluation metrics include both efficiency (i.e., network parameters, floating point operations per second (FLOPs), and frame per second (FPS)) and effectiveness (i.e., Top-1/Top-5 accuracy). As shown in Table~\ref{classification}, our method achieves the best performance nearly in all experimental settings.

\subsection{Object Detection on MS COCO}
Besides the classification task on ImageNet, we also evaluate our method on object detection task to verify its effectiveness and generalization ability. We use our FcaNet with FPN~\cite{lin2017feature} as the backbone (ResNet-50 and ResNet-101) of Faster R-CNN and Mask R-CNN and test their performance on the MS COCO dataset. SENet, CBAM, GCNet, and ECANet are used for comparison.

As shown in Table~\ref{detection}, our method could also achieve the best performance with both Faster-RCNN and Mask-RCNN framework.
Identical to the classification task on ImageNet, FcaNet could also outperform SENet by a large margin with the same number of parameters and computational cost. Compared with the SOTA method ECANet, FcaNet could outperform it by 0.9-1.3\% in terms of AP.

\subsection{Instance Segmentation on MS COCO}
Besides the object detection, we then test our method on the instance segmentation task. As shown in Table~\ref{segmentation}, our method outperforms other methods by a more considerable margin. Specifically, FcaNet outperforms GCNet by 0.5\% AP, while the gaps between other methods are roughly 0.1-0.2\%. These results verify the effectiveness of our method.



\begin{table}[h]
	\centering
	\caption{Instance segmentation results of different methods using Mask R-CNN on COCO val 2017.}
	\label{segmentation}
	\begin{tabular}{lccc}
		\toprule
		Method & AP & $AP_{50}$ & $AP_{75}$ \\
		\hline
		ResNet-50\ \ \   & 34.1 & 55.5 & 36.2 \\
		SENet\ \ \  & 35.4 & 57.4 & 37.8 \\
		GCNet\ \ \  & 35.7 & 58.4 & 37.6 \\
		ECANet\ \ \  & 35.6 & 58.1 & 37.7 \\
		\rowcolor{tb_bg_color}
		FcaNet-LF\ \ \  & \textbf{36.3} & 58.3 & 38.6 \\
		\rowcolor{tb_bg_color}
		FcaNet-TS\ \ \  & \textbf{36.2} & 58.6 & 38.1\\
		\rowcolor{tb_bg_color}
		FcaNet-NAS\ \ \ & \textbf{36.3} & 58.3 & 38.6 \\
		\bottomrule

	\end{tabular}
\end{table}
\section{Conclusion}

In this paper, we study a fundamental problem of channel attention, that is, how to represent channels and regard this problem as a compression process. We have proved that GAP is a special case of DCT and proposed the FcaNet with the multi-spectral attention module, which generalizes the existing channel attention mechanism in the frequency domain. Meanwhile, we have explored different combinations of frequency components in our multi-spectral framework and proposed three criteria for frequency components selection. With the same number of parameters and computational cost, our method could consistently outperform SENet. We also have achieved state-of-the-art performance on image classification, object detection, and instance segmentation compared with other channel attention methods. Moreover, FcaNet is simple yet effective. Our method could be implemented with only a few lines of code change based on existing channel attention methods.

\section*{Acknowledgements}
This work is supported in part by National Key Research and Development Program of China under Grant 2020AAA0107400, Zhejiang Provincial Natural Science Foundation of China under Grant LR19F020004, key scientific technological innovation research project by Ministry of Education, and National Natural Science Foundation of China under Grant U20A20222.

{\small
\bibliographystyle{ieee_fullname}
\bibliography{egbib}
}

\end{document}